\documentclass[10pt,twocolumn,letterpaper]{article}

\usepackage[pagenumbers]{cvpr} 

\usepackage{times}
\usepackage{epsfig}
\usepackage{graphicx}
\usepackage{amsmath}
\usepackage{amssymb}
\usepackage{multirow}
\usepackage{lipsum}
\usepackage{adjustbox}
\usepackage{booktabs}
\usepackage{makecell}
\usepackage{tabu}
\usepackage{bm}
\usepackage{xcolor}
\usepackage{comment}
\usepackage{colortbl}
\usepackage{color}
\usepackage{pifont}
\usepackage{colortbl}
\usepackage{enumitem}
\usepackage{hhline}
\usepackage{mathrsfs}
\usepackage{float}
\usepackage{listings}

\definecolor{Gray}{gray}{0.95}
\definecolor{Grayy}{gray}{0.6}
\definecolor{qingse}{HTML}{177cb0}
\definecolor{juse}{HTML}{ca6924}

\newcommand{\paragrapha}[2][2pt]{\vspace{#1}\noindent\textbf{#2}}

\definecolor{cvprblue}{rgb}{0.21,0.49,0.74}
\usepackage[pagebackref,breaklinks,colorlinks,allcolors=cvprblue]{hyperref}
\def\paperID{xxxx} 
\def\confName{CVPR}
\def\confYear{2025}

\usepackage{pifont}
\title{Ponder \& Press: Advancing Visual GUI Agent \\towards General Computer Control}

\author{
Yiqin Wang$^{*}$ \quad Haoji Zhang$^{*}$ \quad Jingqi Tian \quad Yansong Tang$^{\dag}$  \\
Shenzhen International Graduate School, Tsinghua University \\
\texttt{\{yq-wang23@mails.,tang.yansong@sz.\}tsinghua.edu.cn}\\
}

\begin{document}
\maketitle
\def\thefootnote{$*$}\footnotetext{Equal contribution. \dag Correspondence to Yansong Tang.}

\begin{abstract}
Most existing GUI agents typically depend on non-vision inputs like HTML source code or accessibility trees, limiting their flexibility across diverse software environments and platforms. Current multimodal large language models (MLLMs), which excel at using vision to ground real-world objects, offer a potential alternative. However, they often struggle with accurately localizing GUI elements—a critical requirement for effective GUI automation—due to the semantic gap between real-world objects and GUI elements. In this work, we introduce Ponder \& Press, a divide-and-conquer framework for general computer control using only visual input. Our approach combines an general-purpose MLLM as an `interpreter', responsible for translating high-level user instructions into detailed action descriptions, with a GUI-specific MLLM as a `locator' that precisely locates GUI elements for action placement. By leveraging a purely visual input, our agent offers a versatile, human-like interaction paradigm applicable to a wide range of applications. Ponder \& Press locator outperforms existing models by +22.5\% on the ScreenSpot GUI grounding benchmark. Both offline and interactive agent benchmarks across various GUI environments—including web pages, desktop software, and mobile UIs—demonstrate that Ponder \& Press framework achieves state-of-the-art performance, highlighting the potential of visual GUI agents. Refer to the project homepage \href{https://invinciblewyq.github.io/ponder-press-page/}{here}.
\end{abstract}

\section{Introduction}
\label{sec:intro}

Researchers have long pursued the development of autonomous agents to assist humans in interacting with diverse GUI devices~\cite{world_of_bits, yao2022webshop, li2020mapping}.
With recent advancements in Large Language Models (LLMs)~\cite{brown2020language, OpenAI2023GPT4TR, touvron2023llama, claude35},
agents for web browsing~\cite{gur2024a}, office automation~\cite{wu2024copilot, tan2024cradle},
and mobile apps~\cite{rawles2024androidinthewild} have been proposed to streamline user interactions
and enhance productivity.
Major technology companies have also contributed to this development by creating agents that facilitate user experiences,
such as Apple Siri, Microsoft 365 Copilot, and Capcut smart video editor.

Despite these advancements, existing GUI automation approaches face limitations in generalizability and adaptability across software environments.
First, software-specific agents from tech companies often operate beneath the GUI layer, bypassing user-facing elements and thus sacrificing generalization by interacting directly with underlying code.
Second, most GUI agents~\cite{world_of_bits, humphreys2022data, gur2024a, yao2022webshop, li2020mapping, zhou2023webarena, deng2024mind2web}
developed by the research community rely on supplementary information such as HTML, DOM, or accessibility trees,
which makes them specific to certain platforms and software environments.
Human interaction with GUIs, by contrast, relies exclusively on visual input and interaction through actions
like mouse clicks, keyboard input, and screen taps. \textbf{Therefore, a robust GUI agent designed for broad applicability shall ideally be able to operate using only visual input, similar to human perception, and output actions in a human-like manner.}

Developing vision-only general agents capable of human-like interactions with GUIs presents significant challenges. These challenges can be summarized as:

\begin{itemize}
    \item \textbf{Task Decomposition:} Interpreting and breaking down high-level task instructions into a series of executable actions within a software environment, ensuring that the GUI agent executes the correct action.
    \item \textbf{Precise GUI Localization:} Accurately localizing GUI elements to facilitate the correct placement of actions, such as clicks or text inputs.
\end{itemize}

\begin{figure*}[t]
    \centering
    \begin{minipage}[t]{0.525\linewidth}
        \centering
        \includegraphics[width=\linewidth]{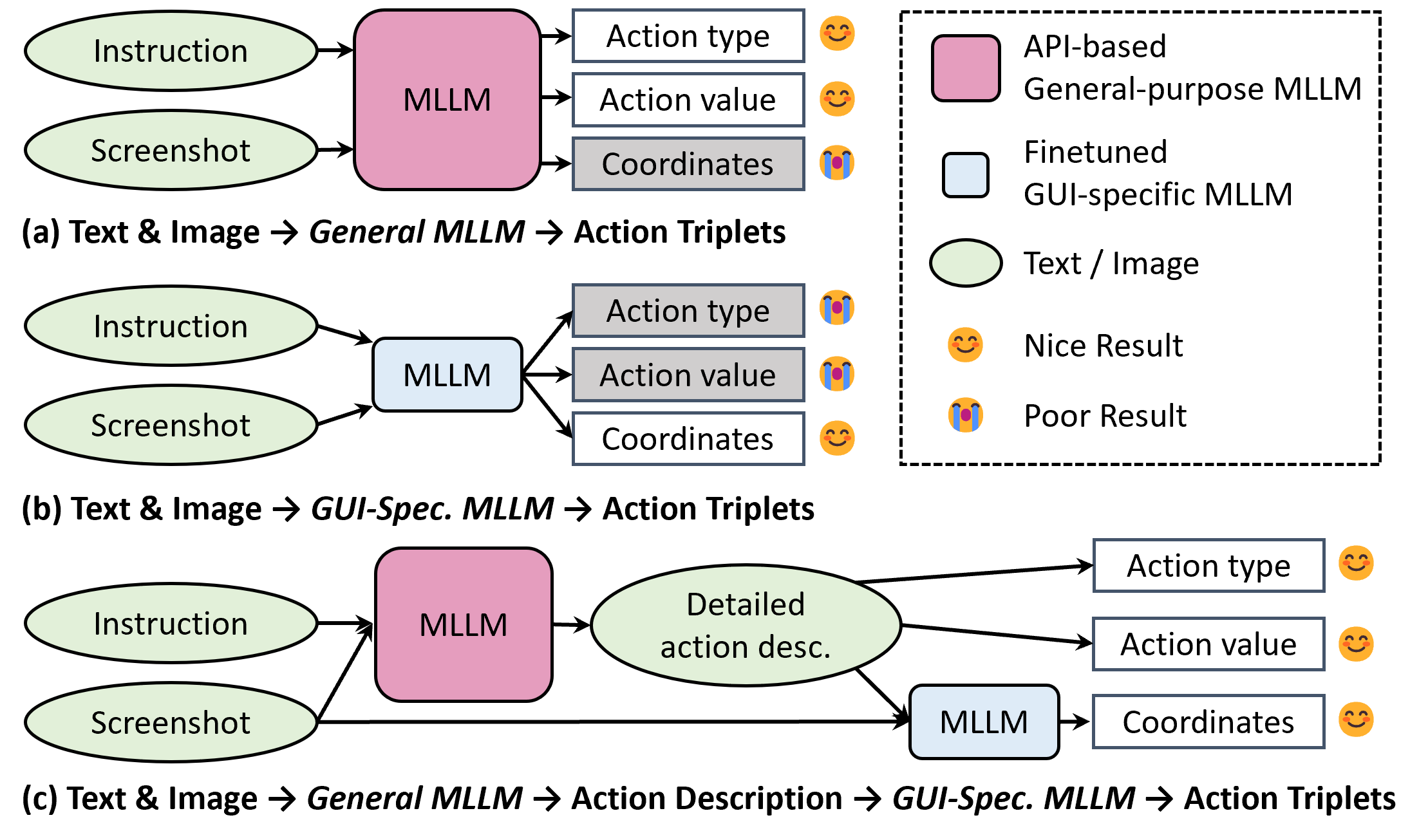}
        \caption{\textbf{Different types of frameworks for vision-based GUI agents.}}
        \label{fig:system}
    \end{minipage}
    \hspace{0.03\linewidth} 
    \begin{minipage}[t]{0.425\linewidth}
        \centering
        \includegraphics[width=\linewidth]{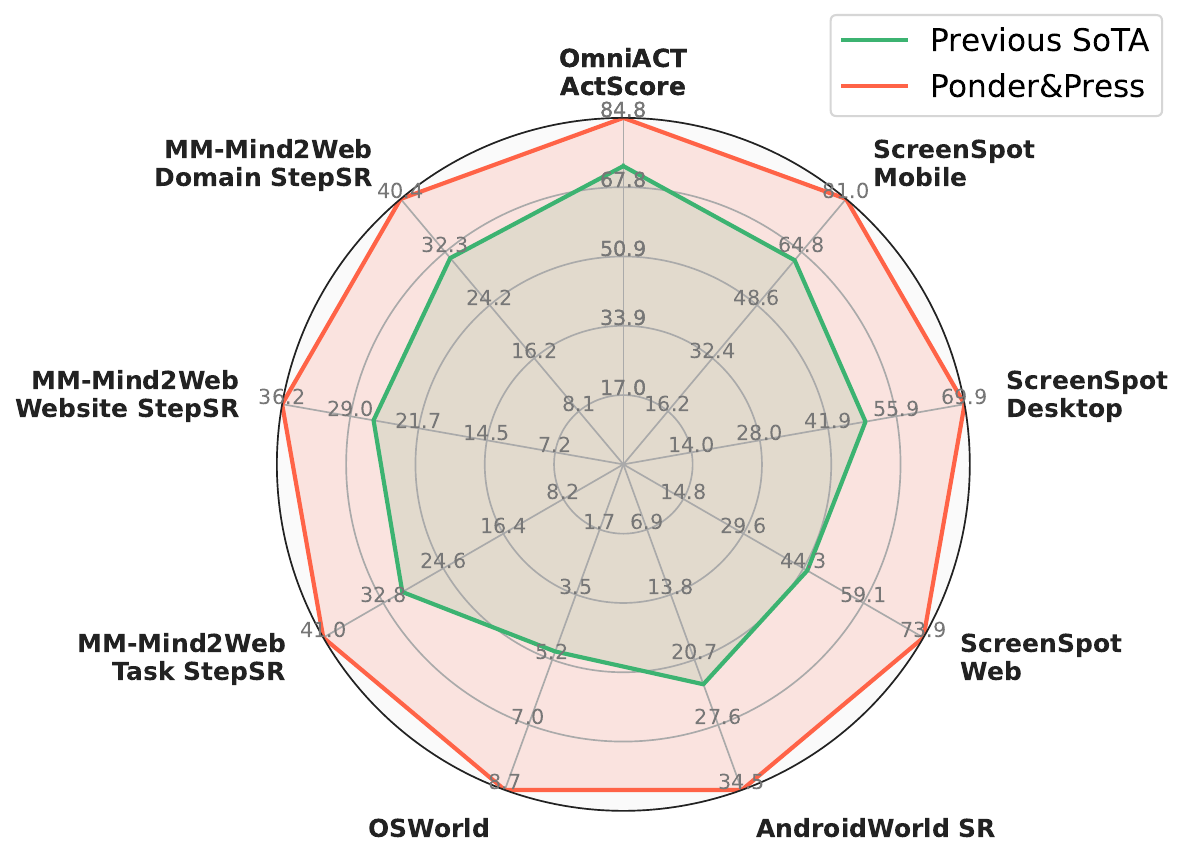}
        \caption{\textbf{\emph{Ponder\&Press} improves vision-based GUI agent on a broad range of tasks.}}
        \label{fig:sota}
    \end{minipage}
\end{figure*}

As shown in ~\Cref{fig:system} (a) and (b), previous efforts have sought to build end-to-end models that address both challenges simultaneously. High-level user instructions are directly mapped to action types, action values, and pixel coordinates in a single inference. However, this approach struggles due to the significant difference between the textual nature of actions and values, and the numerical nature of pixel coordinates. As shown in ~\Cref{fig:system} (a), general-purpose end-to-end multimodal models (MLLMs)~\cite{OpenAI2023GPT4TR, claude35, wang2024qwen2} often suffer from poor grounding performance. As shown in ~\Cref{fig:system} (b), GUI-specific models~\cite{cheng2024seeclick}, though specialized to GUI grounding, struggle with effectively decomposing complex user instructions. As a result, this types of models suffer from poor accuracy in predicting the action type (e.g. TYPE or CLICK) and action value (e.g. the typed content). The claims made above are proved in the experiment section.

In this paper, we introduce a divide-and-conquer framework called Ponder \& Press. It follows the design presented in ~\Cref{fig:system} (c), leveraging the user-instruction interpretation ability of general purpose MLLM, as well as the grounding ability of GUI-specific MLLM. 

\begin{figure*}[t]
    \centering
    \includegraphics[width=0.9\linewidth]{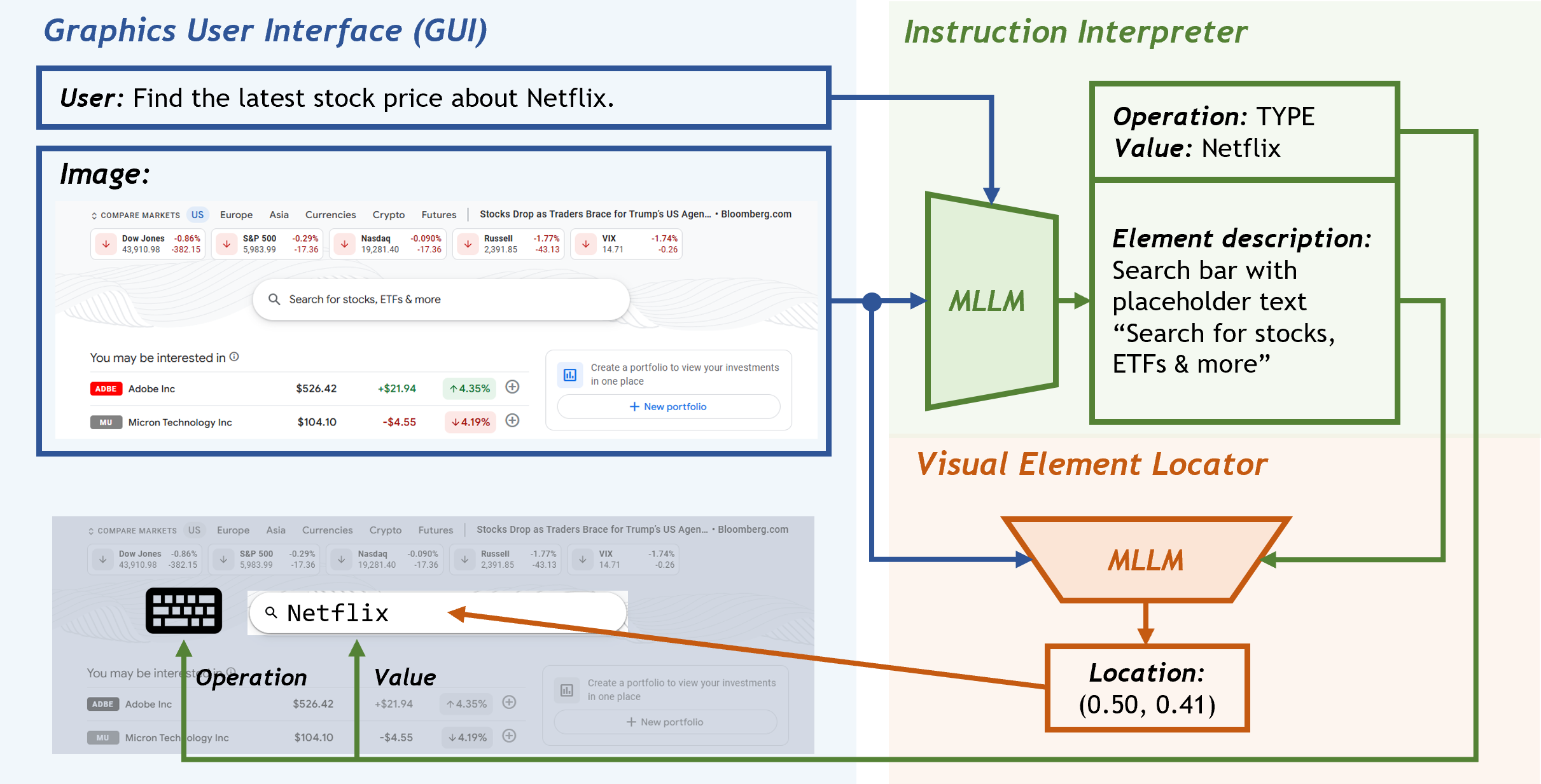}
    \caption{\textbf{The framework of \emph{Ponder\&Press} agent. } The framework consists of two core components:
        an \textcolor{ForestGreen}{\textbf{Instruction Interpreter}} that translates high-level user instructions into actionable steps,
        and a \textcolor{juse}{\textbf{Visual Element Locator}} that localizes GUI elements for interactions such as clicking or typing. Our method ensures that complex instructions can be decomposed and precisely executed within diverse GUIs.}
    \label{fig:teaser}
\end{figure*}

As further shown in ~\Cref{fig:teaser}, the framework is composed of two distinct stages that dealt with the two challenges separately: \textcolor{ForestGreen}{\textbf{(1) The `Ponder' stage}}, involves an\textcolor{ForestGreen}{\textbf{ Instruction Interpreter}} that converts high-level user goals into executable steps. For instance, as shown in ~\Cref{fig:teaser}, when tasked with finding the stock price of `Netflix' on Google Finance, the interpreter outputs: `To find the latest price of Netflix stock, I need to search for Netflix in the Google Finance platform. The search bar is visible at the top of the page, so I'll use that to enter [Netflix]`, along with a structured output "Action: TYPE, Value: `Netflix', Element Description: `Search bar with placeholder text [Search for stocks, ETFs \& more]'". This stage leverages the commonsense knowledge embedded in MLLMs to bridge the gap between high-level user instructions and textual, structured action descriptions. \textcolor{juse}{\textbf{(2) The `Press' stage}}, where we train a \textcolor{juse}{\textbf{Visual Element Locator}} to map the `Element Description' to pixel coordinates, requiring only a small labeled dataset while achieving state-of-the-art performance.

This modular design allows the agent to accurately understand user intent and execute precise actions~\cite{liu2025plan}, maintaining flexibility for general software control. Furthermore, by relying solely on visual inputs—without the need for HTML, accessibility trees, or other supplementary data—our purely visual GUI agent enhances generalizability across various platforms, avoiding the need for software-specific modifications.

Our main contributions are as follows:
\begin{itemize}[left=1em]
    \item We propose Ponder \& Press, a divide-and-conquer GUI agent framework that only relies on visual input to mimic human-like interaction with GUIs. It guarantees the generalizability across diverse environments.
    \item We evaluate Ponder \& Press locator on the GUI grounding benchmark \textit{ScreenSpot}, outperforming previous state-of-the-art model by +22.5\%.
    \item We further conducted extensive evaluations of our framework on 4 widely used GUI agent benchmarks, demonstrating the effectiveness of our agent in offline, online, desktop, webpage, and mobile settings.
\end{itemize}

\section{Related Work}

\subsection{Autonomous Agents for GUI Devices}

System-specific agents developed by technology companies are typically integrated beneath the GUI layer and interact with the underlying code. While providing a highly optimized user experience and being effective within specific environments, these agents are not adaptable to other GUIs, as they lack a mechanism to generalize to arbitrary software interfaces without internal system access. In contrast, many GUI agents developed by the research community, including WebShop~\cite{yao2022webshop}, RCI Agent~\cite{kim2023RCI}, WebArena~\cite{zhou2023webarena}, and MindAct~\cite{deng2024mind2web}, are designed to work with various GUIs but often rely on HTML, DOM, or accessibility trees as input sources to locate elements. This reliance on non-visual data sources limits their applicability to platforms where such data is available, as well as their ability to generalize to GUIs that do not expose internal structural data.

Efforts to build human-like vision-only GUI agents aim to overcome these limitations by focusing solely on visual inputs~\cite{shaw2023pixels, hong2024cogagent, cheng2024seeclick}. However, existing vision-only agents often face challenges with task decomposition and localization precision. For example, single-step end-to-end models that predict both actions and pixel coordinates in a single inference tend to struggle with the modality gap between action description and numerical coordinates, leading to restricted planning ability~\cite{cheng2024seeclick}. Our work addresses these issues by adopting a divide-and-conquer approach to separate task planning from localization, enhancing both the generalizability and precision of our GUI agent.

\subsection{Multimodal Large Language Models}

With the advent of Large Language Models (LLMs)\cite{OpenAI2023GPT4TR, touvron2023llama}, multimodal extensions (MLLMs) have gained traction, enabling new capabilities in processing both text and images. These general-purpose MLLMs, such as GPT-4 series\cite{OpenAI2023GPT4TR} and Claude 3.5 series\cite{claude35}, excel in commonsense reasoning and high-level planning, making them suitable for interpreting complex instructions within a GUI context. MLLMs are able to effectively bridge the gap between high-level user goals into structured actions, leveraging vast amounts of pre-trained knowledge.

Open-source MLLMs such as LLaVA~\cite{liu2023llava, liu2024improved}, Qwen2-VL~\cite{wang2024qwen2}, and their variant~\cite{zhang2024narrative, ye2024voco, zhang2024flash} are designed to solve various vision-related tasks. These models are particularly effective when applied to familiar visual domains that match their training distribution, such as grounding real-world objects. However, their performance declines in out-of-distribution (OOD) scenarios, such as novel GUI layouts, due to limited training data and a narrow generalization capacity. 

In our approach, we leverage the instruction interpretation ability of general-purpose MLLMs for task decomposition while addressing GUI localization with a dedicated visual grounding model, leveraging the commonsense GUI knowledge as well as GUI-specific grounding ability.

\subsection{Visual Grounding}

Visual grounding, the task of associating textual descriptions with specific regions in an image, is foundational to GUI element localization. Early approaches of closed-set object detection are typically trained with IoU-based loss~\cite{FocalLoss}. Recent visual grounding and referring segmentation methods have sought to generalize visual grounding by treating numerical outputs as natural language, which allows for a broader, flexible, and unified approach across various vision-related tasks~\cite{Su2020VL-BERT, wang2024qwen2, bai2024self, liu2024universal, liu2024open, han2023global, yang2024language}.

However, due to the semantic gap between real-world image and GUI image, these general visual grounding models suffer from severe performance drop on GUI data, as further shown in our experiment section. Screenshot marks~\cite{yang2023setofmark, liu2024coarse} and chain-of-thought methods~\cite{wei2022chain, lu2023thinkbot} serve as training-free workarounds to help MLLM understand the relative position between different visual elements. Still, it relies on explicitly performing another stage of image segmentation or object tracking, which is inconvenient and may involve additional error. 

In order to enhance grounding performance on GUI data, ~\cite{UIBert, qian-etal-2024-visual, li-etal-2020-mapping, li-etal-2020-widget} build datasets to bridge the gap between natural language, GUI element, and its location. CogAgent~\cite{hong2024cogagent} conducted large-scale pretraining on datasets including 400k webpage screenshots, and further finetuned on human-annotated restricted internal dataset. SeeClick~\cite{cheng2024seeclick} open-sourced a GUI visual grounding training set consisting of 1M data, from which we sampled a 25K subset to finetune our model.

Our approach builds on these advances by training a GUI-specific grounding model using a small labeled dataset to translate structured action descriptors from our planning stage into precise pixel coordinates. This modular design facilitates robust and efficient GUI localization across diverse environments, ensuring both high accuracy and consistency in action execution.

\section{Method}

\subsection{Task Formulation}

Consider a GUI environment $\mathcal{E}$ (e.g., an office software, a web page, a mobile app interface, etc.) and a task $\mathcal{T}$ (e.g., `Find the latest news about Netflix stock.'). The agent's goal is to produce a sequence of executable actions $\mathcal{A} = [\alpha_1, \alpha_2, \ldots, \alpha_m]$ to complete the task. At each step $k$, the agent $\rho$ must generate an action $\alpha_k$ based on the current visual observation $o_k$, previous actions $\{\alpha_1, \alpha_2, \ldots, \alpha_{k-1}\}$, and the task $\mathcal{T}$:

$$
\alpha_k = \rho(o_k, \mathcal{T}, \{\alpha_1, \alpha_2, \ldots, \alpha_{k-1}\})
$$

In this setting, the observation $o_k$ is purely visual. We have $o_k = i_k$ at each step $k$, with $i_k$ representing the screenshot input. No structured HTML code, DOM tree, accessibility tree or any other text-based information is available. All environment understanding must be derived from the current screenshot. The state of the GUI environment $\mathcal{E}$ updates after each action as follows:

$$
o_{k+1} = \mathcal{E}(\alpha_k)
$$

Each action $\alpha$ corresponds to an application or system event within the environment, represented as a triplet:
$$
\alpha = (\eta, \omega, \nu)
$$

Here, $\eta$ represents a target location (e.g., `[0.50, 0.20]') as a pixel coordinate on the screen, denoting the position where `Click', `Type', or `Select' operation should be executed. 
$\omega \in \mathcal{O}$ specifies the intended operation type (e.g., `Type'), and $\nu$ provides any additional value required for the action (e.g., the type content `netflix'). The set $\mathcal{O}$ encompasses all allowable operations in $\mathcal{E}$.

\subsection{Framework Design}


It is challenging for multimodal language models (MLLMs) to produce the action triplet $(\eta, \omega, \nu)$ in a single inference step. Specifically, generating $\omega$ and $\nu$ requires strong planning abilities, contextual reasoning, and domain-specific knowledge of the GUI, while determining $\eta$ demands precise and accurate grounding of GUI elements. As shown in the experiments section, existing end-to-end models exhibit relatively low performance on this challenging task. To address this issue, we introduce an intermediate variable $\mathcal{D}$, a textual description of the target element that serves as a reliable and interpretable bridge for accurate grounding.

Our approach follows a two-stage process:

1. \textbf{Instruction Interpretation}: The first model $\phi$ functions as an instruction interpreter. Given the current screenshot $o_k$, previous actions $\{\alpha_1, \alpha_2, \ldots, \alpha_{k-1}\}$, and the task $\mathcal{T}$, it generates the intermediate output $(\mathcal{D}, \omega, \nu)$:
   $$
   (\mathcal{D}, \omega, \nu) = \phi(o_k, \mathcal{T}, \{\alpha_1, \alpha_2, \ldots, \alpha_{k-1}\}),
   $$
   where $\phi$ encapsulates task interpretation abilities.

2. \textbf{GUI Element Localization}: The second model $\psi$ functions as a visual GUI element locator. Given the current screenshot $o_k$ and the textual description $\mathcal{D}$, it determines the relative coordinates $\eta$ on the screen:
   $$
   \eta = \psi(o_k, \mathcal{D}),
   $$
   where $\psi$ represents the grounding function for locating GUI elements.

The final action triplet $(\eta, \omega, \nu)$, thus obtained, can be executed within the GUI environment to get the next observation $o_{k+1}$.

\subsection{Instruction Interpreter}

The Instruction Interpreter translates high-level task instructions into structured components for GUI interaction, producing $(\mathcal{D}, \omega, \nu)$ in a single inference. Here, $\mathcal{D}$ is a textual description of the target element, $\omega$ denotes the intended operation (e.g., Click, Type), and $\nu$ provides additional input required for the action, such as specific text or dates.

We employ two multimodal models—GPT-4o and Claude 3.5 Sonnet—that process screenshots alongside the task $\mathcal{T}$ and prior actions $\{\alpha_1, \alpha_2, \ldots, \alpha_{k-1}\}$. Given these inputs, each model generates a text output containing $(\mathcal{D}, \omega, \nu)$:
$$
(\mathcal{D}, \omega, \nu) = \phi(o_k, \mathcal{T}, \{\alpha_1, \alpha_2, \ldots, \alpha_{k-1}\}),
$$
where $\phi$ represents the instruction interpreter. Each output is extracted directly from the model’s single-text response.

\begin{table*}[t]
    \centering
    \caption{\textbf{Comparisons with pure-vision methods on GUI grounding benchmark \textit{ScreenSpot}~\cite{cheng2024seeclick}.} I/W denote Icon/Widget. Results with * are from the SeeClick paper~\cite{cheng2024seeclick}. Ponder \& Press's locator exhibits state-of-the-art performance on precisely locating GUI elements while maintaining a smaller model size. } 
    \adjustbox{width=\linewidth}{
        \begin{tabular}{cccccccccc}
        \hline \multirow{2}{*}{ Methods } & Model & GUI & \multicolumn{2}{c}{ Mobile } & \multicolumn{2}{c}{ Desktop } & \multicolumn{2}{c}{ Web } & \multirow{2}{*}{ Avg. } \\
        \cline { 4 - 9 } & Size & Specific & Text & I/W & Text & I/W & Text & I/W & \\
        \hline Qwen2-VL~\cite{wang2024qwen2} & 7 B & $\boldsymbol{X}$ & $41.4 \%$ & $16.2 \%$ & $25.3 \%$ & $5.7 \%$ & $12.2 \%$ & $6.3 \%$ & $17.8 \%$ \\
        GPT-4o~\cite{OpenAI2023GPT4TR} & N/A & $\boldsymbol{X}$ & $23.4 \%$ & $25.8 \%$ & $17.5 \%$ & $21.4 \%$ & $10.9 \%$ & $9.7 \%$ & $18.1 \%$ \\
        Claude 3.5 Sonnet~\cite{claude35} & N/A & $\boldsymbol{X}$ & $37.6 \%$ & $26.1 \%$ & $29.0 \%$ & $26.3 \%$ & $17.4 \%$ & $8.4 \%$ & $24.1 \%$ \\
        \hline Fuyu*~\cite{fuyu-8b} & 8 B & $\checkmark$ & $40.6 \%$ & $1.6 \%$ & $33.6 \%$ & $6.7 \%$ & $48.4 \%$ & $2.9 \%$ & $22.3 \%$ \\
        CogAgent*~\cite{hong2024cogagent} & 18 B & $\checkmark$ & $66.5 \%$ & $26.7 \%$ & $7 3 . 7 \%$ & $19.3 \%$ & $7 8 . 0 \%$ & $2 1 . 4 \%$ & $47.6 \%$ \\
        SeeClick*~\cite{cheng2024seeclick} & 9.6 B & $\checkmark$ & $7 5 . 9 \%$ & $4 8 . 7 \%$ & $72.7 \%$ & $2 6 . 4 \%$ & $69.2 \%$ & $2 1 . 4 \%$ & $5 2 . 4 \%$ \\
        \rowcolor{Gray} \textbf{Ponder\&Press locator} & 7 B & $\checkmark$ & $\mathbf{88.6} \%$ & $\mathbf{73.4} \%$ & $\mathbf{80.4} \%$ & $\mathbf{59.3} \%$ & $\mathbf{82.6} \%$ & $\mathbf{65.1} \%$ & $\mathbf{74.9} \%$ \\
        \hline
        \end{tabular}
    }
    \label{tab:screenspot}%
\end{table*}%

\subsection{Visual Element Locator}

The Visual Element Locator module is tasked with accurately identifying and locating GUI elements within a screenshot, positioning this as a GUI visual grounding task. The objective is to produce the normalized coordinates $(x, y)$ of the target element, with values constrained to $0 \leq x, y \leq 1$.

For this purpose, we use Qwen2-VL-Instruct~\cite{wang2024qwen2} as the pretrained model and further finetune it with LoRA~\cite{hu2021lora} on a GUI-specific data subset sampled from~\cite{cheng2024seeclick}. This finetuning enhances the model's capacity to localize GUI elements effectively across diverse interfaces.

The Locator computes the coordinates $\eta = (x, y)$ based on the current screenshot observation $o_k$ and the textual description $\mathcal{D}$ generated by the Instruction Interpreter, using the function $\psi$:

$$
\eta = \psi(o_k, \mathcal{D}).
$$

To train the model to output these coordinates, we avoid explicit numerical loss. Instead, we treat the prediction as a natural language next-token-prediction task. We prompt the model following the prompt template presented in~\cite{cheng2024seeclick} as follows:

\begin{quote}
\textit{"In this UI screenshot, what is the position of the element corresponding to the description \{DESCRIPTION \} (with point)?"}
\end{quote}

This setup encourages the model to generate $(x, y)$ as part of a structured textual response, effectively supporting GUI-specific localization in a multimodal environment.

\subsection{Training Details}

Ponder \& Press locator is based on the Qwen2-VL~\cite{wang2024qwen2} model, leveraging its initial multimodal grounding capabilities. In order to fit its output space to GUI data, we apply LoRA~\cite{hu2021lora} to adapt both the visual encoder and language model layers using a 2.5\% subset of the SeeClick training set~\cite{cheng2024seeclick}, totally 25,000 samples.

Training is conducted for 400 steps on 8 NVIDIA A100 GPUs, consuming approximately 2 hours. We use the AdamW optimizer with a learning rate of $3 \times 10^{-5}$ and apply a cosine annealing scheduler to manage learning rate decay. To enhance visual resolution for precise element localization, we employ the Naive Dynamic Resolusion~\cite{wang2024qwen2, NaViT} to extend the input resolution to $896 \times 896$. It enables resolution scaling without additional retraining, which is advantageous for GUI tasks requiring high visual detail.

This simple configuration allows Ponder \& Press to achieve effective grounding across diverse GUI environments, optimizing for both computational efficiency and accuracy in element localization.

\section{Experiments}

We evaluate Ponder \& Press in two main areas: grounding accuracy and agent performance. First, we assess our Visual Element Locator’s grounding ability on the ScreenSpot~\cite{cheng2024seeclick} benchmark to measure precise GUI element localization. Then, we evaluate the entire agent on four GUI agent benchmarks. For offline scenarios, we use Multimodal-Mind2Web~\cite{deng2024mind2web} and OmniACT~\cite{kapoor2024omniact}, and for interactive settings, we test on OSWorld~\cite{xie2024osworld} and AndroidWorld~\cite{rawles2024androidworld}. These experiments demonstrate the effectiveness of our approach across diverse GUI environments.

\begin{table*}[t]
    \centering
    \caption{\textbf{Comparisons with pure-vision methods on web agent benchmark \textit{Multimodal-Mind2Web}~\cite{deng2024mind2web}, in a zero-shot manner.} Claude denote Claude 3.5 Sonnet~\cite{claude35}, Naive Guess denote always ground on the center point of the screen, Ele.Acc denote element accuracy, Step SR denote step success rate. Results with * are from the SeeClick paper~\cite{cheng2024seeclick}. Ponder \& Press exhibits state-of-the-art performance on both element accuracy and step success rate. } 
    \adjustbox{width=\linewidth}{
        \begin{tabular}{cccccccccc}
        \hline Visual & Instruction & GUI & \multicolumn{2}{c}{ Cross-Task } & \multicolumn{2}{c}{ Cross-Website } & \multicolumn{2}{c}{ Cross-Domain } & Avg. \\
        \cline { 4 - 9 } Locator & Interpreter & Specific & Ele.Acc & Step SR & Ele.Acc &  Step SR & Ele.Acc & Step SR & Step SR \\
        \hline Naive Guess & Claude & $\boldsymbol{X}$ & $0.6 \%$ & $0.5 \%$ & $1.6 \%$ & $1.4 \%$ & $1.2 \%$ & $0.9 \%$ & $0.9 \%$\\
        Qwen2-VL~\cite{wang2024qwen2} & Claude & $\boldsymbol{X}$ & $9.1 \%$ & $8.4 \%$ & $11.2 \%$ & $9.7 \%$ & $8.7 \%$ & $7.8 \%$ & $8.6 \%$\\
        \hline SeeClick*~\cite{cheng2024seeclick} & w/o & $\checkmark$ & $26.3 \%$ & $23.7 \%$ & $21.9 \%$ & $18.8 \%$ & $22.1 \%$ & $20.2 \%$ & $20.9 \%$\\
        SeeClick & Claude & $\checkmark$ & $34.9 \%$ & $30.2 \%$ & $32.9 \%$ & $26.5 \%$ & $36.1 \%$ & $31.4 \%$ & $29.4 \%$\\
        \rowcolor{Gray} \textbf{Ponder\&Press} & Claude & $\checkmark$ & $\mathbf{46.7} \%$ & $\mathbf{41.0} \%$ & $\mathbf{44.1} \%$ & $\mathbf{36.2} \%$ & $\mathbf{47.0} \%$ & $\mathbf{40.4} \%$ & $\mathbf{39.2} \%$\\
        \hline
        \end{tabular}
    }
    \label{tab:mind2web}%
\end{table*}%

\subsection{GUI Grounding Benchmark}

\paragrapha{ScreenSpot.} To assess the grounding capabilities of Ponder \& Press's Visual Element Locator, we evaluate it on the ScreenSpot dataset~\cite{cheng2024seeclick}, a benchmark specifically designed for GUI element localization. ScreenSpot encompasses over 600 diverse screenshots from mobile (iOS, Android), desktop (macOS, Windows), and web platforms, along with more than 1,200 instructions tied to actionable elements (as shown in Figure 3, left). This dataset provides a comprehensive test for our model across varied GUI environments. Notably, ScreenSpot includes a significant number of icons and widgets, which are more challenging to locate than standard text elements due to the subtle visual cues required for precise identification.

In ~\Cref{tab:screenspot}, we present a comparison of Ponder \& Press with prior pure-vision models on the ScreenSpot benchmark. Comparing non-GUI-specific models (such as Qwen2-VL and GPT-4o) with those trained on GUI-specific data reveals a distinct advantage for the latter. Non-GUI-specific models struggle with localization due to their lack of targeted knowledge about GUI structures and common interface patterns. In contrast, our fine-tuning equips the model with essential prior knowledge, enhancing its ability to navigate and precisely localize elements in varied and complex GUIs. 

As shown in ~\Cref{tab:screenspot}, Ponder\&Press locator model achieves state-of-the-art performance across all GUI categories—mobile, desktop, and web—outperforming previous methods by a substantial margin. In particular, Ponder \& Press surpasses the previous SoTA, SeeClick~\cite{cheng2024seeclick}, with an average accuracy increase of over 20\%. Our model demonstrates particular strength in locating icon and widget elements, underscoring the model’s robust learning of such GUI-specific visual features.

\subsection{Offline GUI Agent Benchmark}

\begin{table}[t]
    \centering
    \caption{\textbf{Comparisons with pure-vision methods on desktop and web agent benchmark \textit{OmniACT}~\cite{kapoor2024omniact}.} Seq. denote sequence, Act. denote action, Claude denote Claude 3.5 Sonnet~\cite{claude35}, Naive Guess denote always ground on the center point of the screen. Ponder \& Press's exhibits state-of-the-art performance on action score, proving its strong GUI grounding capability. } 
    \begin{tabular}{ccccc}
    \hline Visual & Instruction & Seq. & Act. & Click \\
    Locator & Interpreter & Score & Score & Penalty\\
    \hhline{|=====|} \multirow{2}{*}{Naive Guess} & GPT-4o & $30.9$ & $55.8$ & $36.7$ \\
    & Claude & $39.3$ & $58.8$ & $32.8$ \\
    \hline \multirow{2}{*}{QWen2-VL} & GPT-4o & $30.9$ & $70.4$ & $22.1$\\
    & Claude & $39.3$ & $70.0$ & $21.5$\\
    \hline \multirow{2}{*}{SeeClick} & GPT-4o & $30.9$ & $73.0$ & $19.5$ \\
    & Claude & $39.3$ & $73.1$ & $18.4$ \\
    \hline  \multirow{2}{*}{\textbf{Ponder\&Press}} & GPT-4o & $30.9$ & $\mathbf{84.8}$ & $\mathbf{7.7}$ \\
    & Claude & $39.3$ & $\mathbf{82.9}$ & $\mathbf{8.6}$ \\
    \hline
    \end{tabular}
    \label{tab:omniact}%
\end{table}%

\paragrapha{Multimodal Mind2Web~\cite{deng2024mind2web}.} We first evaluate the Ponder \& Press agent on the Multimodal-Mind2Web benchmark in a \textbf{zero-shot} manner. The benchmark includes three test splits designed to assess generalization across various dimensions: \textbf{(1) Cross-Domain}, which assesses the agent’s ability to generalize to new high-level domains (Information and Service), including 912 tasks across 73 websites; \textbf{(2) Cross-Website}, which evaluates generalization to new websites within familiar domains, including 177 tasks from 10 websites per domain; and \textbf{(3) Cross-Task}, which evaluates the agent's ability to address new tasks, including 252 tasks across 69 websites.

As shown in ~\Cref{tab:mind2web}, we focus on two core evaluation metrics: \textbf{element accuracy (Ele.Acc)} and \textbf{step success rate (Step SR)}, omitting operation F1 as a primary metric. While operation F1 measures action type correctness, simply setting every operation as `CLICK' yields high F1 scores across all splits (over 80\%), but fails to capture the nuance required for varied interaction types, resulting in poor step success rates. Thus, operation F1 metric is included only in supplementary materials to avoid misleading conclusions about the agent’s effectiveness.


\begin{table*}[t]
    \centering
    \caption{\textbf{Comparisons with visual methods on online GUI agent benchmark \textit{OSWorld}~\cite{xie2024osworld}.} Results with * are from OSWorld~\cite{xie2024osworld}. Ponder \& Press exhibits unified performance boost across all subsets comparing to the GPT-4o baseline.}
        \begin{tabular}{cccccccc|c}
        \hline \multicolumn{3}{c}{\multirow{2}{*}{Methods}} & Office & OS & Daily & Workflow & Professional & All \\
        &&& (117) & (24) & (78) & (101) & (49) & (369)\\
        \hhline{|=========|}
        \multicolumn{3}{c}{Human*} & $71.8$ & $75.0$ & $70.5$ & $73.3$ & $73.5$ & $72.4$ \\
        \hhline{|=========|} \rowcolor{Gray} &&& \multicolumn{6}{c}{Single-stage method} \\
        \multicolumn{3}{c}{CogAgent*} & $0.9$ & $4.2$ & $2.7$ & $0.0$ & $0.0$ & $1.1$ \\
        \multicolumn{3}{c}{GPT-4o*} & $3.6$ & $8.3$ & $6.1$ & $5.6$ & $4.1$ & $5.0$ \\
        \hhline{|=========|} \rowcolor{Gray} Agent & Locator & Interpreter & \multicolumn{6}{c}{Two-stage method} \\
        Ponder\&Press & Naive Guess & GPT-4o & $0.0$ & $0.0$ & $0.0$ & $0.0$ & $0.0$ & $0.0$ \\
        Ponder\&Press & SeeClick & GPT-4o & $5.1$ & $16.7$ & $7.7$ & $5.0$ & $6.1$ & $6.5$ \\
        \textbf{Ponder\&Press} &\textbf{Ponder\&Press} & \textbf{GPT-4o} & $\mathbf{6.8}$ & $\mathbf{16.7}$ & $\mathbf{12.8}$ & $\mathbf{5.0}$ & $\mathbf{10.2}$ & $\mathbf{8.7}$ \\
        \hline
        \end{tabular}
    \label{tab:osworld}%
\end{table*}%

\begin{table}[t]
    \centering
    \caption{\textbf{Comparisons with methods on interactive mobile GUI agent benchmark \textit{AndroidWorld}~\cite{rawles2024androidworld}. } \textbf{P\&P denote Ponder\&Press.} Results with * are from AndroidWorld~\cite{rawles2024androidworld}. With only visual input, Ponder \& Press agent, equipped with the visual locator, exhibits state-of-the-art performance on Success Rates.}
        \begin{tabular}{cccc}
        \hline \multirow{2}{*}{Agent} & Visual & Instruction &  Success\\
        & Locator & Interpreter &  Rates\\
        \hhline{|====|} Human* & - & - & $80 \%$ \\
        \hhline{|====|} \rowcolor{Gray} \multicolumn{4}{c}{Input: A11y tree} \\
        M3A*~\cite{rawles2024androidworld} & GPT-4 Turbo & GPT-4 Turbo & $30.6 \%$ \\
        \hhline{|====|} \rowcolor{Gray} \multicolumn{4}{c}{Input: Screenshot + A11y tree} \\
        M3A*~\cite{rawles2024androidworld} & GPT-4 Turbo & GPT-4 Turbo & $25.4 \%$ \\
        \hhline{|====|} \rowcolor{Gray} \multicolumn{4}{c}{Input: Screenshot} \\
        P\&P & Naive Guess & GPT-4o & $0.0 \%$ \\
        P\&P & SeeClick & GPT-4o & $23.3 \%$ \\
        \textbf{P\&P} & \textbf{P\&P} & \textbf{GPT-4o} & $\mathbf{34.5} \%$ \\
        \hline
        \end{tabular}
    \label{tab:androidworld}%
\end{table}%


\Cref{tab:mind2web} illustrates that our Ponder \& Press model, equipped with the Claude 3.5 Sonnet interpreter, achieves state-of-the-art performance on both element accuracy and step success rate across all three splits. With the same interpreter, the locator of Ponder \& Press demonstrates a substantial improvement ($+9.8 \%$) compared to the previous SoTA SeeClick\cite{cheng2024seeclick}. Moreover, when we equip SeeClick's original model with our Claude interpreter, the model also demonstrates a significant boost in performance ($+8.5 \%$). This underscores the generalizability of our interpretation stage, which enhances GUI-specific grounding accuracy and task success across diverse and dynamic web contexts. Intuitively, our locator model outperforms non-GUI-specific models (such as Qwen2-VL) by a significant margin ($+30.6 \%$), confirming the effectiveness of grounding capabilities honed on GUI-specific data. This also clearly proves that a strong locator is necessary to fully unleash the potential of the instruction interpreter.

\paragrapha{OmniACT\cite{kapoor2024omniact}.} We further evaluate the Ponder \& Press agent on the OmniACT benchmark~\cite{kapoor2024omniact}, which consists of over 9.8K pairs of images and instructions from various operating systems and the web. The dataset includes screenshots of diverse UI screens, paired with corresponding natural language instructions, and the goal is to generate executable commands using the PyAutoGUI Python library. OmniACT employs two key evaluation metrics: \textbf{sequence score} and \textbf{action score}. The sequence score measures whether the predicted action sequence matches the ground truth, while the action score evaluates how well the generated code snippet performs the task. Note that the sequence score is only impacted by the accuracy of the action sequence produced, making it an independent metric from the action score and a measure of the planning ability.

As shown in \Cref{tab:omniact}, our results demonstrate that Ponder \& Press achieves state-of-the-art performance in terms of the action score. The notable increase in action score is primarily driven by a significant reduction in click penalties. For example, when equipped with GPT-4o as the interpreter, Ponder \& Press demonstrates a $11.8\%$ increase in Action Score while simultaneously decreasing exactly $11.8\%$ in Click Penalty, compared to the previous SoTA, SeeClick\cite{cheng2024seeclick}. This remarkable phenomenon confirms that our model excels at precisely locating the specific GUI element that needs to be clicked—an essential component of effective grounding. Furthermore, we observe that the Claude 3.5 Sonnet interpreter consistently outperforms GPT-4o in both action prediction and element description, which underscores the critical importance of adopting a strong interpreter for effective task execution.

It is worth noting that the evaluation script open-sourced by OmniACT does not fully align with the formula provided in the Section 4 of their paper, where the action score should only be calculated for those action sequences that match. Instead, their script calculates the action score across all samples, including those where the sequence is mismatched. To avoid confusion and ensure fair comparison, we do not directly compare our results with those reported in the OmniACT paper as those results are much lower than ours due to the mis-implementation of the formula. Our independent evaluation, following the correct action score formula, highlights the superior grounding and planning abilities of Ponder \& Press. Please refer to supplementary materials for detailed discussion.

\subsection{Interactive Online GUI Agent Benchmark}

\paragrapha{OSWorld~\cite{xie2024osworld}.} OSWorld is a computer environment designed to evaluate multimodal GUI agents in a execution-based manner. OSWorld proposed a benchmark of 369 computer tasks, involving OS-related tasks, office-related tasks, as well as desktop apps operation tasks such as Chrome, VLC Player, and Thunderbird. As shown in ~\Cref{tab:osworld}, Ponder\&Press outperforms GPT-4o baseline in all task categories, with notable performance gains in office-related, OS-related, and daily tasks. This unified improvement across the benchmark solidifies the effectiveness of our agent in handling a wide range of GUI-related tasks. Furthermore, agent equips with Ponder\&Press locator surpasses the one with SeeClick locator, which proves the precise localization capabilities of Ponder\&Press and aligns the results on visual grounding benchmark. When equipping our agent with the SeeClick locator, it still surpasses the GPT-4o baseline, further proving the generalizability and effectiveness of our divide-and-conquer framework.

\paragrapha{AndroidWorld~\cite{rawles2024androidworld}.} AndroidWorld is a dynamic Android environment that evaluates interactive mobile GUI agents across 116 tasks from 20 real-world apps. It generates tasks expressed in natural language, offering a flexible and realistic testing suite. As shown in ~\Cref{tab:androidworld}, Ponder\& Press agent equipped with the Ponder\&Press visual locator achieves a success rate of 34.5\%, a notable improvement over the SeeClick locator variant (23.3\%). This result validate the effectiveness of our locator model. Even when compared to the strong baseline M3A which utilizes A11y trees input, our agent demonstrate a superior performance, proving the effectiveness of our vision-only framework.

\subsection{Case study}

\begin{figure}[t]
    \centering
    \includegraphics[width=1.0\linewidth]{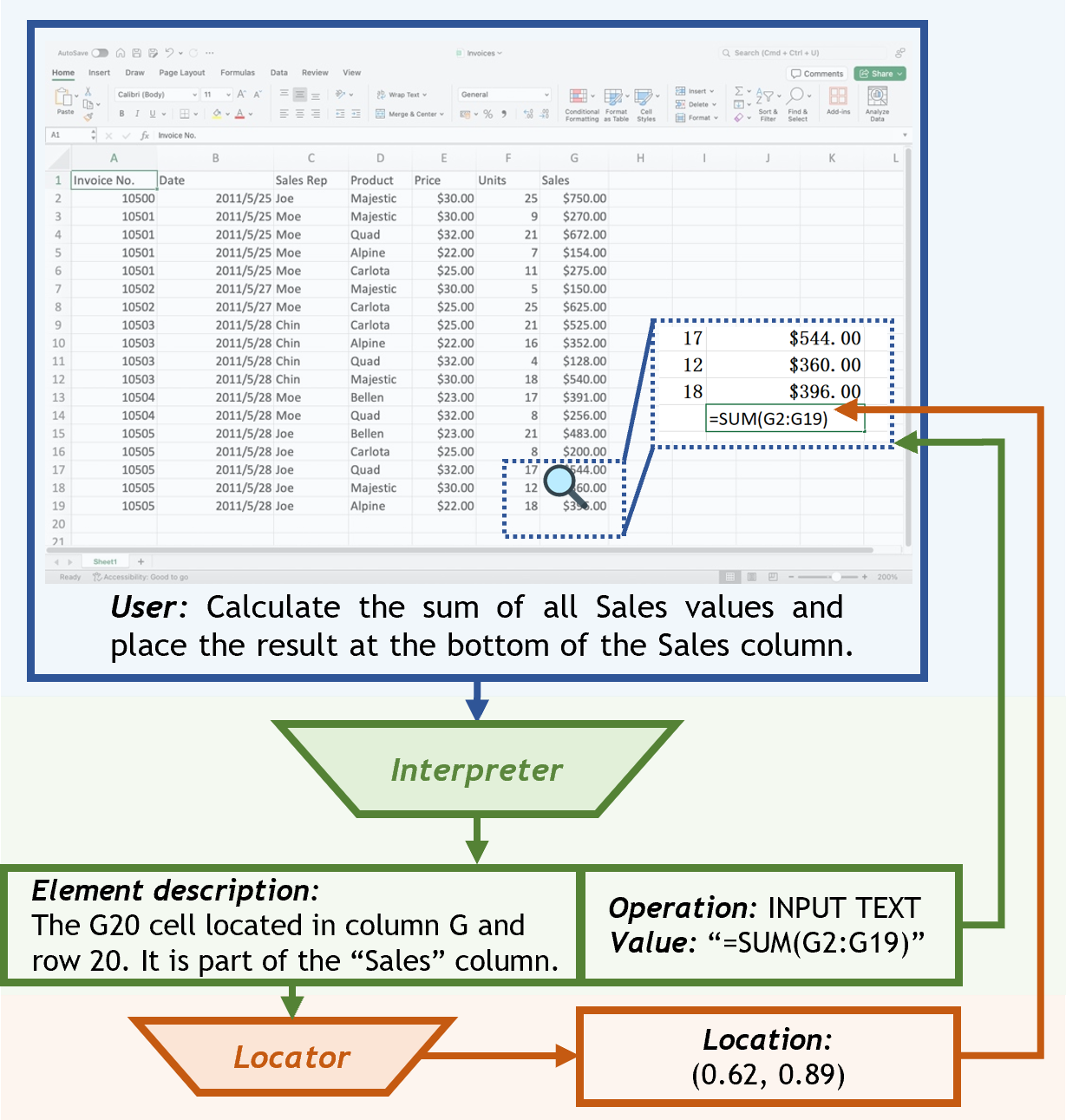}
    \caption{\textbf{Case study of Ponder\&Press on an office task.}}
    \label{fig:case}
\end{figure}

To better demonstrate the effectiveness of Ponder\&Press, we hereby provide a case study.
As shown in ~\Cref{fig:case}, the user needs to calculate the sum of a specific row in Microsoft Excel. The interpreter determined a `INPUT TEXT' operation is needed, and the typed value should be a SUM expression in Excel. The interpreter also output the description of the GUI element, at which the operation should be conducted. Given this description, the locator output the central coordinate of that GUI element, denoting a specific cell in Excel sheet. With the correct action triplet, our agent successfully finished the proposed task. For cases on more platforms and software environments, please refer to the supplementary material.

\section{Future Work.}

\begin{figure}[t]
    \centering
    \includegraphics[width=1.0\linewidth]{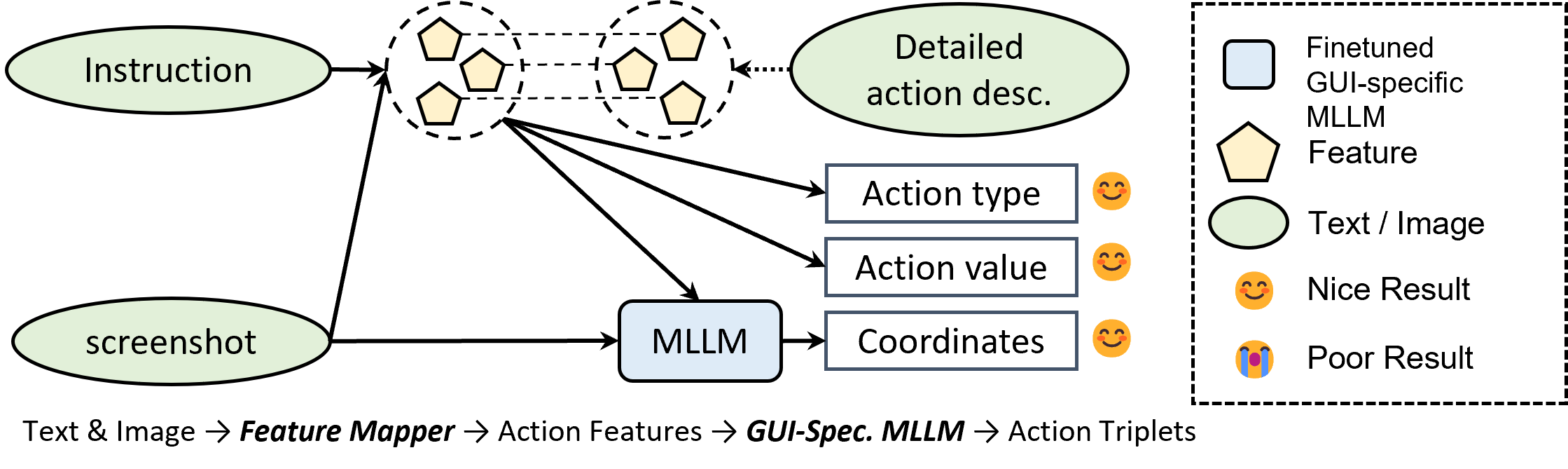}
    \caption{\textbf{A possible end-to-end visual GUI agent framework.}}
    \label{fig:future}
\end{figure}

Just like any other agents that involve API-based model~\cite{rawles2024androidworld, tan2024cradle}, our agent inevitably involves inference latency and API cost. Previous single-stage methods suffer from either poorly mapping high-level user instruction into action description~\cite{cheng2024seeclick, hong2024cogagent}, or poorly predicting the coordinates for action placement~\cite{OpenAI2023GPT4TR, claude35}. To resolve these problems while not involving API-based model, we further propose a possible end-to-end visual GUI agent framework as future work. 

As shown in ~\Cref{fig:future}, we could align the features between (high-level instruction + screenshot) and (detailed action description) in the first training stage, and train the GUI-specific MLLM to enhance grounding ability in the second training stage. This framework could possibly encompass a strong GUI locating capability while retaining the task decomposition ability. Main challenge may lies in the collection of large-scale, high-quality training data.

\section{Conclusion}

We introduced Ponder\&Press, a divide-and-conquer framework that enables general computer control using only visual input. The framework consists of two stages: the `Ponder' stage, which interprets high-level instructions into actionable steps, and the `Press' stage, which accurately localizes GUI elements for action placement. By relying solely on visual input, Ponder\&Press ensures generalizability across diverse software environments without needing supplementary data like HTML or accessibility trees. Through extensive offline and online evaluations across desktop, web, and mobile environments, Ponder\&Press achieved state-of-the-art performance, demonstrating both precise localization and effective task decomposition. Our work showcases the potential of vision-based GUI agents for general-purpose automation, paving the way for flexible, human-like interactions with diverse software systems.

{
    \small
    \bibliographystyle{ieeenat_fullname}
    \bibliography{main}
}
\clearpage

\appendix

\section*{Appendix}

\section{Evaluation Details}

\subsection{ScreenSpot}

~\paragrapha{Deal with redundant output.} ScreenSpot~\cite{cheng2024seeclick} evaluates a model's visual grounding performance based on its predicted relative coordinates $(x,y)$, where $0 \leq x,y \leq 1$. For models not fine-tuned on GUI-specific data~\cite{OpenAI2023GPT4TR, claude35, wang2024qwen2}, their grounding outputs often include redundant descriptive text alongside the $(x,y)$ coordinates. To evaluate these models on ScreenSpot and accurately report their GUI grounding performance, we use Regular Expression (Regex) to extract the coordinates while discarding any extraneous information. The following Python function performs this coordinate extraction:

\begin{lstlisting}[language=Python]
def extract_two_float_tuple(s):
    pattern = r'[\(\[\s]*([-+]?\d*\.\d+|\d+)\s*,\s*([-+]?\d*\.\d+|\d+)\s*[\)\]\s]*|-\s*[Xx]:\s*([-+]?\d*\.\d+|\d+)\s*(?:\([^\)]*\))?\s*-\s*[Yy]:\s*([-+]?\d*\.\d+|\d+)\s*(?:\([^\)]*\))?|-\s*[Tt]op:\s*([-+]?\d*\.\d+|\d+)\s*-\s*[Ll]eft:\s*([-+]?\d*\.\d+)|\(\s*[Xx]:\s*([-+]?\d*\.\d+|\d+)\s*,\s*[Yy]:\s*([-+]?\d*\.\d+|\d+)\s*\)'
    match = re.search(pattern, s, re.VERBOSE)
    if match:
        if match.group(1) and match.group(2):
            return float(match.group(1)), float(match.group(2))
        elif match.group(3) and match.group(4):
            return float(match.group(3)), float(match.group(4))
        elif match.group(5) and match.group(6):
            return float(match.group(6)), float(match.group(5))
        else:
            return float(match.group(8)), float(match.group(9))
    else:
        raise ValueError("String does not contain a valid '(float, float)', '[float, float]', '- X: float - Y: float', '- X: float (xxxxxx) - Y: float (xxxxxx)', '- Top: float - Left: float', or '(X: float, Y: float)' pattern")
\end{lstlisting}

In cases where the function raises an error due to no match, we fall back to the default result of $(0.5,0.5)$, representing the center point. As tested, over 95\% of cases match successfully. Consequently, the ScreenSpot results we report for non GUI specific models~\cite{OpenAI2023GPT4TR, claude35, wang2024qwen2} in main paper Table.1 reliably reflect their GUI grounding performance.

\subsection{Multimodal-Mind2Web}

\begin{table*}[t]
    \centering
    \caption{\textbf{Comparisons with pure-vision methods on web agent benchmark \textit{Multimodal-Mind2Web}~\cite{deng2024mind2web}.} Claude denote Claude 3.5 Sonnet~\cite{claude35}, Ele.Acc denote element accuracy, Op.F1 denote operation F1, Step SR denote step success rate. \textbf{Always conducting `CLICK' action may result in higher operation F1, but harms step success rate.}} 
    \adjustbox{width=\linewidth}{
        \begin{tabular}{ccc|cccccccccc}
        \hline Visual & Ele. Desc. & Action & \multicolumn{3}{c}{ Cross-Task } & \multicolumn{3}{c}{ Cross-Website } & \multicolumn{3}{c}{ Cross-Domain } \\
        \cline { 4 - 12 } Locator & Interpreter & Interpreter & Ele.Acc & Op.F1 & Step SR & Ele.Acc & Op.F1 & Step SR & Ele.Acc & Op.F1 & Step SR \\
        \hline SeeClick & GPT-4o & GPT-4o & $31.7 \%$ & $72.6 \%$ & $28.3 \%$ & $33.0 \%$ & $72.3 \%$ & $27.2 \%$ & $34.3 \%$ & $71.6 \%$ & $30.6 \%$ \\
        \rowcolor{Gray} \textbf{Ponder\&Press} & GPT-4o & Always CLICK & $\mathbf{42.8} \%$ & $\mathbf{83.5} \%$ & $32.7 \%$ & $\mathbf{43.8} \%$ & $\mathbf{80.8}\%$  & $32.2 \%$ & $\mathbf{45.3} \%$ & $\mathbf{83.8} \%$ & $35.5 \%$ \\
        \rowcolor{Gray} \textbf{Ponder\&Press} & GPT-4o & GPT-4o & $\mathbf{42.8} \%$ & $72.6 \%$ & $\mathbf{37.0} \%$ & $\mathbf{43.8} \%$ & $72.3\%$  & $\mathbf{36.8} \%$ & $\mathbf{45.3} \%$ & $71.6 \%$ & $\mathbf{39.4} \%$ \\
        \hline SeeClick & Claude & Claude & $34.9 \%$ & $79.2 \%$ & $30.2 \%$ & $32.9 \%$ & $76.2 \%$ & $26.5 \%$ & $36.1 \%$ & $79.0 \%$ & $31.4 \%$ \\
        \rowcolor{Gray} \textbf{Ponder\&Press} & Claude & Always CLICK & $\mathbf{46.7} \%$ & $\mathbf{83.5} \%$ & $35.7 \%$ & $\mathbf{44.1} \%$ & $\mathbf{80.8}\%$  & $30.7 \%$ & $\mathbf{47.0} \%$ & $\mathbf{83.8} \%$ & $35.1 \%$ \\
        \rowcolor{Gray} \textbf{Ponder\&Press} & Claude & Claude & $\mathbf{46.7} \%$ & $79.2 \%$ & $\mathbf{41.0} \%$ & $\mathbf{44.1} \%$ & $76.2\%$  & $\mathbf{36.2} \%$ & $\mathbf{47.0} \%$ & $79.0 \%$ & $\mathbf{40.4} \%$ \\
        \hline
        \end{tabular}
    }
    \label{tab:mind2web_spp}%
\end{table*}%

\paragrapha{Discussion on metrics.} Mind2Web adopt 3 metrics: (1) Element Accuracy measures whether the selected GUI element is one of the acceptable elements, reflecting the grounding accuracy. (2) Operation F1 calculates token-level F1 score for the predicted action. For those action type without value (such as `CLICK'), this is the same as action type accuracy. For those action type with value (such as `TYPE' needs a content), this is calculated between the predicted value and GT value. (3) Step Success Rate is measured among all steps, a step is regarded as successful only if both the selected element and the predicted operation are correct.

In our agent setting, action type is fully determined by the instruction interpreter. As shown in the first, the third, the forth, and the sixth line of supplementary~\Cref{tab:mind2web_spp}.
However, due to the highly biased action type distribution in Mind2Web, simply setting every operation as `CLICK' yields a high F1 scores across all splits (over 80\%) as shown in the second and fifth line of supplementary~\Cref{tab:mind2web_spp}. This brute force manner results in failure of all `non-CLICK' steps, harming the step success rate. This reflects that simply comparing the Op.F1 metric is not enough to determine the action predicting ability. We claim that the gold metric should always be the Step Accuracy, which relies on the correctness of both element selection and action prediction.

\subsection{OmniACT}

\paragrapha{Calculation details of Action Score.} According to the formula presented in OmniACT~\cite{kapoor2024omniact} and shown below, when $\text{SeqScore}_i = 0$, the i-th case does not contribute to the final result of the Action Score. In other words, the Action Score should only be calculated based on matched sequences. Moreover, the sum of (Action Score + Click Penalty + Key Penalty + Write Penalty) is always expected to equal 100\%, and mismatched sequences should not introduce penalties.
\begin{align*}
&\textbf{Action Score} \\
=&
\frac{
\sum_i 
\max \left(
\text{SeqScore}_i 
- \sum_j\left(M_i^j + K_i^j + W_i^j\right), 
0
\right)
}{
\sum_i \text{SeqScore}_i
}
\end{align*}

However, in the evaluation script provided by OmniACT, mismatched sequences still incur penalties. Specifically, the condition $\text{SeqScore}_i \geq 1$ is not properly enforced in the code. As a result, (Action Score + Click Penalty + Key Penalty + Write Penalty) equals $\text{SeqScore}$ instead of 100\%, as reflected in the experimental results table of OmniACT~\cite{kapoor2024omniact}. This implementation error leads to a misreported Action Score that fails to accurately represent "how well a code snippet performs the correct action." The reported values are directly influenced by the $\text{SeqScore}$ due to this mistake. 

To address this issue, we report all results based on the correct formula, ensuring consistency with the intended evaluation framework.

\subsection{OSWorld}
When building Ponder\& Press agent on OSWorld~\cite{xie2024osworld} benchmark, we firstly prompt the interpreter model to explicitly generate 1.action type, 2.action value, and 3.GUI element description. Then we utilize the locator model to convert GUI element description into pixel-level coordinates. With the action triplet (action type, action value, coordinates), we finally format the `pyautogui' code required by OSWorld in a rule-based manner.

\subsection{AndroidWorld}
When building Ponder\& Press agent on AndroidWorld benchmark, we refer to the prompts of M3A agent proposed in the AndroidWorld paper~\cite{rawles2024androidworld}. The main difference lies in we solely utilize raw screenshot as input, neither utilizing set-of-mark labels~\cite{yang2023set} nor utilizing additional a11y tree input. We use the output coordinate of our grounding model to decide action placement location on the screen.

\subsection{Model Endpoints}
We utilize api-based MLLM as the instruction interpreter. To better ensure consistent behaviors, we listed the model endpoint names as follows:

\begin{itemize}
    \item GPT-4o: `gpt-4o-2024-08-06'
    \item Claude: `aws\_claude35\_sdk\_sonnet'
\end{itemize}

\section{More examples}

\begin{figure*}[t]
    \centering
    \includegraphics[width=0.75\linewidth]{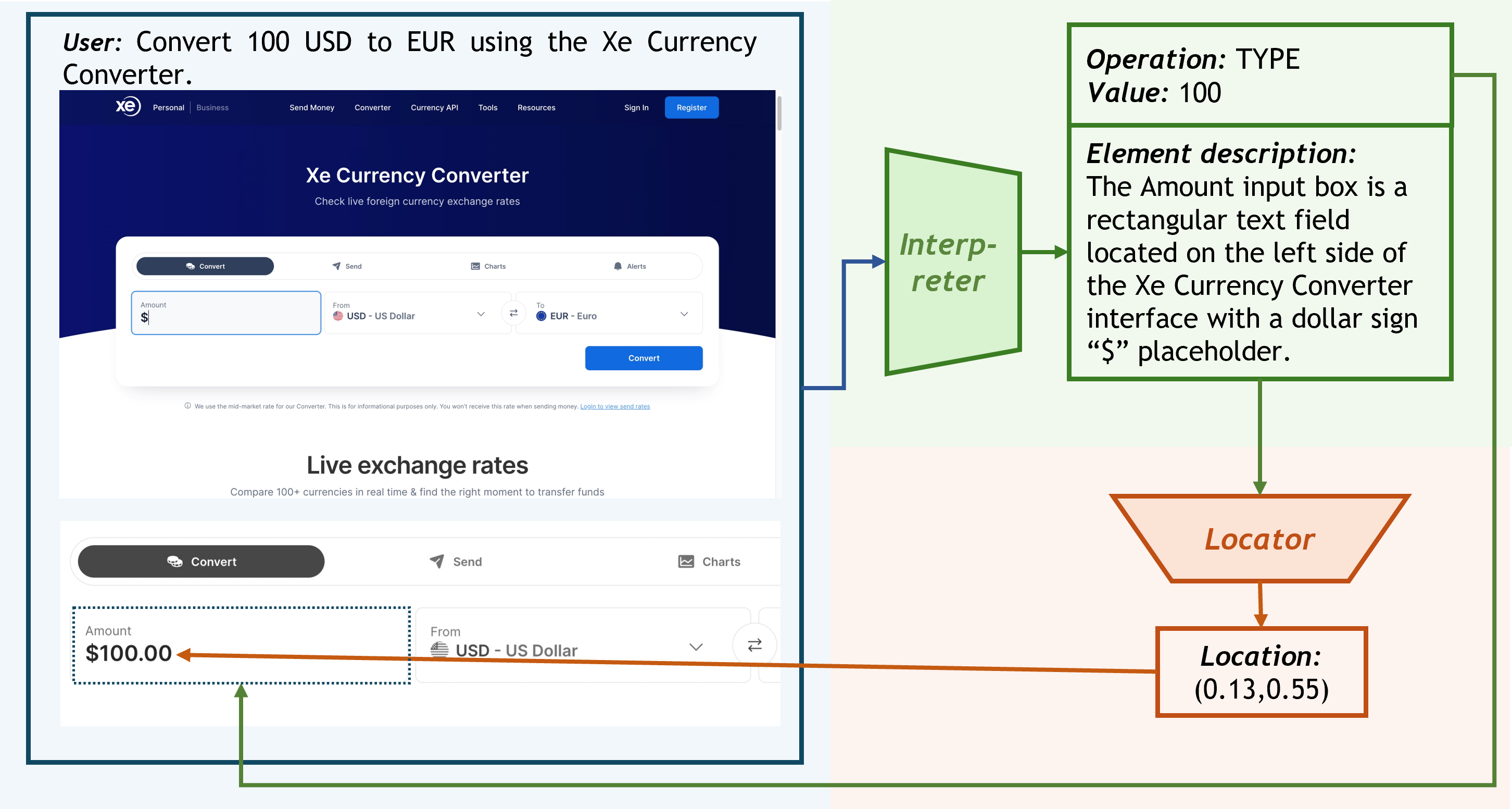}
    \caption{Example of webpage GUI task.}
    \label{fig:supp3}
\end{figure*}

\begin{figure*}[t]
    \centering
    \includegraphics[width=0.75\linewidth]{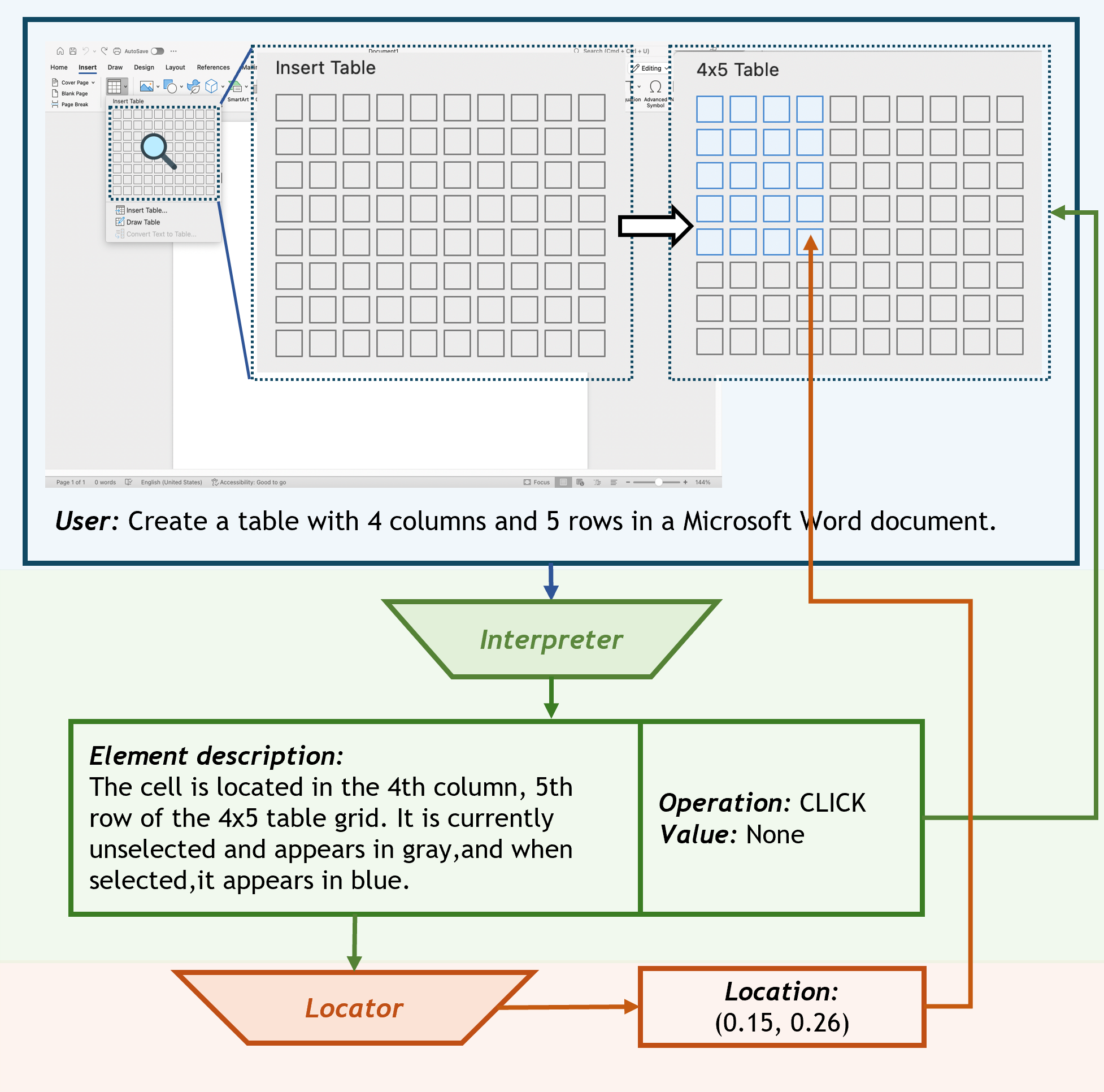}
    \caption{Example of office GUI task.}
    \label{fig:supp4}
\end{figure*}

\begin{figure*}[t]
    \centering
    \includegraphics[width=0.53\linewidth]{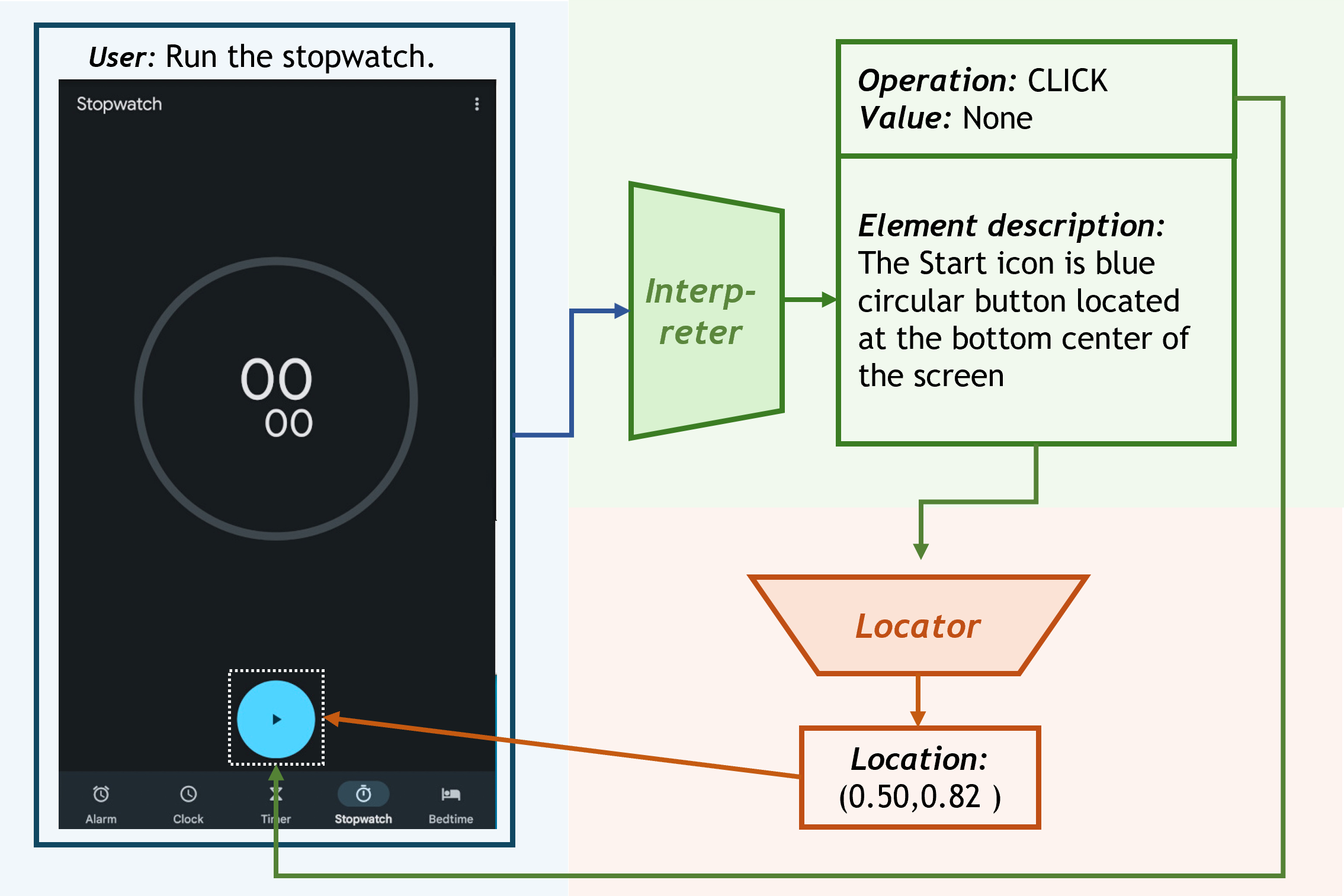}
    \caption{Example of mobile GUI task.}
    \label{fig:supp1}
\end{figure*}

\begin{figure*}[t]
    \centering
    \includegraphics[width=0.53\linewidth]{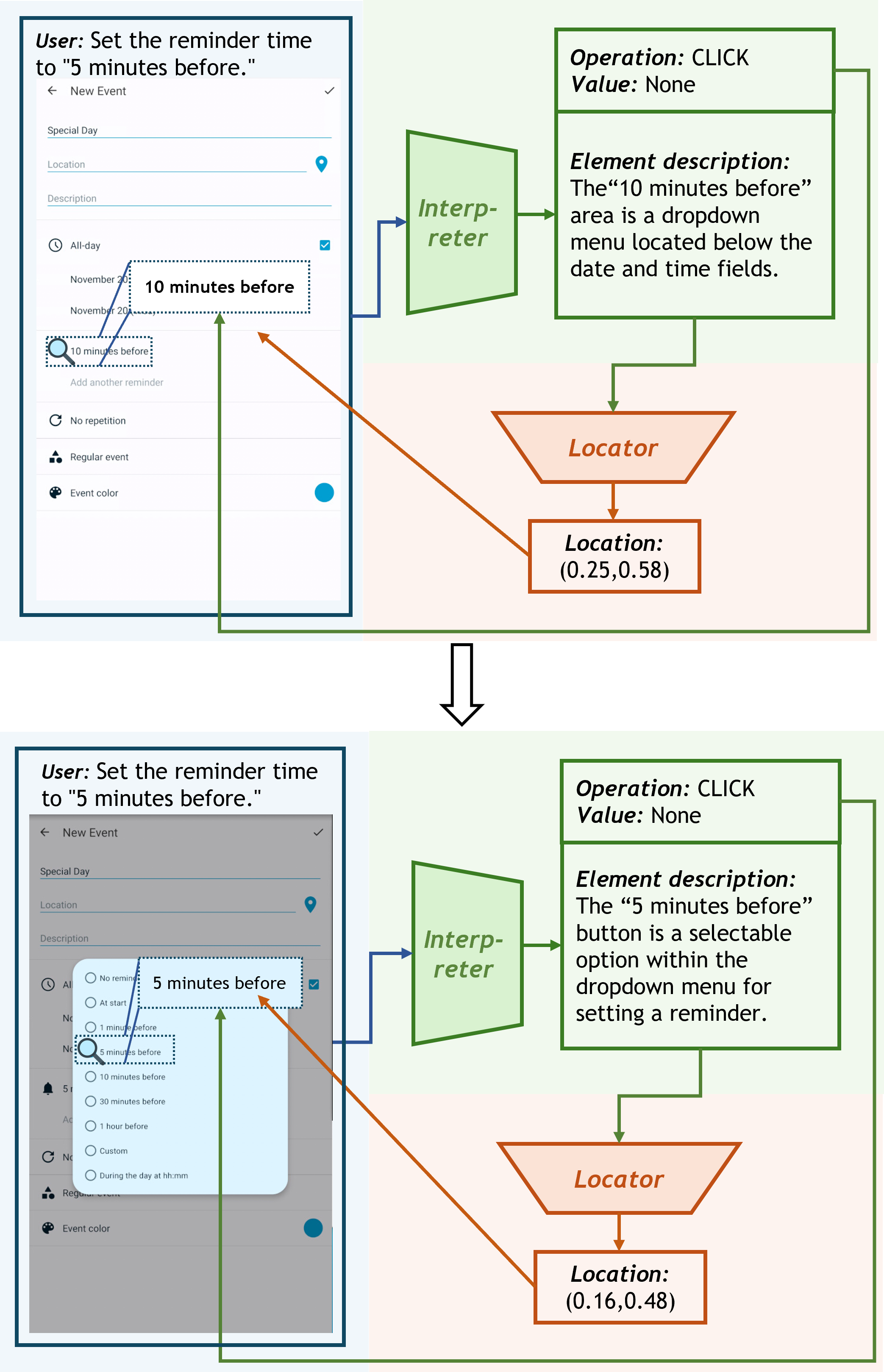}
    \caption{Example of mobile GUI task.}
    \label{fig:supp2}
\end{figure*}

To better showcase the pipeline of Ponder\&Press, we provide more examples at supplementary~\Cref{fig:supp3}, ~\Cref{fig:supp4}, ~\Cref{fig:supp1}, and ~\Cref{fig:supp2}.

\end{document}